\documentclass[letterpaper, 10 pt, conference]{ieeeconf}  

\IEEEoverridecommandlockouts                              

\let\labelindent\relax

\usepackage{graphicx}
\usepackage{amsmath}
\usepackage{amssymb}  
\usepackage{amsfonts}
\usepackage{multirow}
\usepackage{booktabs}
\usepackage{wrapfig}
\usepackage{threeparttable}
\usepackage[inline]{enumitem}
\usepackage{color}

\title{\LARGE \bf
Adaptive Hierarchical SpatioTemporal Network for Traffic Forecasting
}

\author{Yirong Chen$^{1}$, Ziyue Li$^{2}$, Wanli Ouyang$^{3}$, Michael Lepech$^{4}$, 
\thanks{$^{1}$Yirong Chen is with the Department of Civil and Environmental Engineering, Stanford University, California  94305, U.S.A 
         {\tt\small chenyr@stanford.edu <chenyr@stanford.edu>}}%
\thanks{$^{2}$Ziyue Li is with Information System, University of Cologne, 50923 Cologne, NRW, Germany {\tt\small zlibn@wiso.uni-koeln.de}}%
\thanks{$^{3}$Wanli Ouyang is now a professor at Shanghai AI Lab, Shanghai, China {\tt\small wanli.ouyang@sydney.edu.au}}%
\thanks{$^{4}$Michael Lepech is with the Department of Civil and Environmental Engineering, Stanford University, California  94305, U.S.A {\tt\small mlepech@stanford.edu}}%
 }

\begin{document}

\maketitle
\thispagestyle{empty}
\pagestyle{empty}

\begin{abstract}
Accurate traffic forecasting is vital to intelligent transportation systems, which are widely adopted to solve urban traffic issues. Existing traffic forecasting studies focus on modeling spatial-temporal dynamics in traffic data, among which the graph convolution network (GCN) is at the center for exploiting the spatial dependency embedded in the road network graphs. However, these GCN-based methods operate intrinsically on the node level (e.g., road and intersection) only whereas overlooking the spatial hierarchy of the whole city. Nodes such as intersections and road segments can form clusters (e.g., regions), which could also have interactions with each other and share similarities at a higher level. In this work, we propose an Adaptive Hierarchical SpatioTemporal Network (AHSTN) to promote traffic forecasting by exploiting the spatial hierarchy and modeling multi-scale spatial correlations. Apart from the node-level spatiotemporal blocks, AHSTN introduces the adaptive spatiotemporal downsampling module to infer the spatial hierarchy for spatiotemporal modeling at the cluster level. Then, an adaptive spatiotemporal upsampling module is proposed to upsample the cluster-level representations to the node-level and obtain the multi-scale representations for generating predictions. Experiments on two real-world datasets show that AHSTN achieves better performance over several strong baselines.

\end{abstract}
\section{Introduction}
The fast growth of urbanization has posed significant urban traffic problems. To alleviate urban traffic issues, Intelligent Transportation System (ITS) has seen active research and development in the recent decade own to its effectiveness in optimizing transportation system's performance and sustainability. As the cornerstone of a successful ITS, traffic forecasting aims to provide the ITS with an accurate prediction of future traffic states (such as traffic speed, volume, and demand) based on current and historical conditions \cite{wang2023correlated,li2022profile}. An accurate and efficient traffic forecasting system can work with advanced traffic management systems and advanced traveler information systems in ITS to reduce traffic congestion and improve system efficiency, which further benefits user experience, decreases energy consumption, and promotes sustainability \cite{lan2023mm}.

Traditionally, time series models such as Auto-Regressive Integrated Moving Average (ARIMA) and Vector Auto-Regressive (VAR), as well as high-dimensional methods such as tensor decomposition and prediction \cite{li2020tensor,li2020long}, are used for traffic forecasting \cite{kamarianakisForecastingTrafficFlow2003,williamsModelingForecastingVehicular2003}. However, limited by their linear nature, these methods cannot capture the high non-linearity prevalent in traffic conditions, which restrains their forecast ability. With the development of deep learning, neural network (NN)-based approaches are increasingly explored in traffic forecasting studies. Recurrent neural networks such as GRU and LSTM \cite{zhangCombiningWeatherCondition2017,yuDeepLearningGeneric2017,kangShorttermTrafficFlow2017} along with temporal convolution networks \cite{yuSpatioTemporalGraphConvolutional2018} are used to capture the temporal correlations. Convolution neural network (CNN) models are proposed to capture spatial correlations \cite{zhangDeepSpatioTemporalResidual2017,jiangGeospatialDataImages2019}. However, CNN approaches require the traffic network to be modeled as grids, which does not always hold for real-world scenarios where the traffic networks are often non-Euclidean \cite{geng2019spatiotemporal,ziyue2021tensor,li2022individualized}, namely, graph-structured. 

To overcome this problem, the most advanced models have shifted to graph convolutional networks (GCN) for spatial correlations modeling, such as STGCN \cite{yuSpatioTemporalGraphConvolutional2018}, DCRNN \cite{liDiffusionConvolutionalRecurrent2018}, Graph WaveNet \cite{wuGraphWaveNetDeep2019}, AGCRN \cite{baiAdaptiveGraphConvolutional2020}, and TVDBN \cite{lin2023dynamic}, which have shown great improvement in forecasting accuracy. However, these GCN-based works model the complex city-scale traffic involvement on the road network level only, where nodes are intersections or road segments and edges are the connections between the nodes. This approach overlooks the importance of \textit{spatial hierarchy} in the road network. Specifically, intersections and road segments can form clusters or belong to certain regions. The clusters and regions can have spatial interactions or share similarities with each other that need to be captured.  

To utilize the spatial hierarchy of the road network for traffic forecasting, HGCN \cite{guoHierarchicalGraphConvolution2021} first uses spectral clustering to form clusters based on the raw traffic data in an offline manner and then employs separate GCN networks to capture spatial correlations at different scales. While HGCN achieves promising performance, the power of spatial hierarchy in traffic forecasting may still be underestimated since the offline clustering process could be sub-optimal for two reasons. First, the raw traffic data often fluctuates and can be very noisy, which would harm the accuracy of the clustering process. Second, by directly clustering on the raw data rather than on the feature embeddings, HGCN does not consider the spatiotemporal correlations among the traffic data when forming the clusters and thus has a disadvantage in spatiotemporal learning. 



In this work, we propose an \textbf{A}daptive \textbf{H}ierarchical \textbf{S}patio-\textbf{T}emporal \textbf{N}etwork (AHSTN) to infer the spatial hierarchy in the road network together with modeling the spatiotemporal correlations in an end-to-end manner. This enables the model to learn more accurate spatiotemporal correlations from both node-level and cluster-level. Specifically, AHSTN introduces a learnable Adaptive SpatioTemporal Downsampling (AST-D) module to adaptively downsample the original graph by clustering the nodes (e.g., at the road network level) to clusters (e.g., at the region level), based on which spatiotemporal correlations in the cluster level can be captured with spatiotemporal blocks (e.g., the blocks proposed in STGCN). Correspondingly, an Adaptive SpatioTemporal Upsampling (AST-U) module is utilized to upsample the cluster-level spatiotemporal representations to the node level, which will be combined with the node-level representation learned with the original graph to capture the multi-scale and hierarchical spatiotemporal correlations. The learnable parameters used in AST-D and AST-U modules can be optimized with the spatiotemporal blocks in an end-to-end manner to make AHSTN a more unified framework and capture more accurate spatiotemporal correlations. To evaluate the effectiveness of AHSTN, we conduct extensive experiments on two real-world traffic datasets by performing multi-step prediction and demonstrate that AHSTN achieves better performance than a set of strong baselines. 
In summary, our main contributions are as follows: 
\begin{itemize}
    \item We propose an Adaptive Hierarchical Spatio-Temporal Network (AHSTN) for traffic forecasting, which infers the hierarchical structure in the traffic network together with the spatiotemporal network optimization in an end-to-end manner for learning spatiotemporal correlations in multiple scales.
     \item We introduce an Adaptive SpatioTemporal Downsampling module to downsample the original graph for learning cluster-level representations and an Adaptive SpatioTemporal Upsampling module to upsample the cluster-level representations to the node-level.
    \item We conduct extensive experiments on two real-world urban traffic datasets and achieve state-of-the-art performance compared to a set of strong baselines. 
\end{itemize}

\section{Related Work}

\subsection{GCN-based traffic forecasting} 
Application of graph convolutional network (GCN) has seen increasing popularity in traffic forecasting problems \cite{jiangGraphNeuralNetwork2022}, as a graph is the natural representation of a traffic network. For example, STGCN \cite{yuSpatioTemporalGraphConvolutional2018} is the first work to apply purely convolutional structures to extract spatio-temporal features from graph-structured time series data for traffic forecasting problems by proposing ST-Conv blocks, which sandwich a single spatial graph convolution layer between two temporal gated convolution layers. Then, many GCN-based methods are proposed to solve the traffic forecasting problems, such as DCRNN \cite{liDiffusionConvolutionalRecurrent2018} that formulates the spatial relations as a diffusion process and extends the application of GCN to directed graphs, Graph WaveNet \cite{wuGraphWaveNetDeep2019} that proposes a self-adaptive adjacent matrix module and dilated causal convolution networks to uncover unseen graph connections beyond the pre-defined adjacency matrix, AGCRN \cite{baiAdaptiveGraphConvolutional2020} that introduces a node adaptive parameter learning module to generate node-specific features from a shared parameter pool using matrix factorization, HGCN \cite{guoHierarchicalGraphConvolution2021} that adopts spectral clustering to generate spatial hierarchy and uses the attention mechanism to control the relationship between nodes and clusters dynamically, etc. Among these methods, only HGCN considers the hierarchy information of the graph-structured time series data. But they adopt the pre-learned clustering of nodes, which may fail to capture the true dynamics of node-cluster interaction because of the spatiotemporal property of traffic conditions. To alleviate this issue, we propose an end-to-end GCN approach to adaptively learn the clustering on feature embeddings rather than raw traffic data.

\subsection{Graph Clustering} The idea of graph clustering is to create a spatial hierarchy based on the original flat graph \cite{grattarolaUnderstandingPoolingGraph2022}. There are various model-free clustering approaches such as Graclus \cite{dhillonWeightedGraphCuts2007}, Node Decimation Pooling (NDP) \cite{bianchiHierarchicalRepresentationLearning2022}, and Eigen Pooling \cite{maGraphConvolutionalNetworks2019}. 
However, model-free clustering methods do not further optimize the clusters once they are formed, which makes them overlook the dynamic of node-cluster interaction for downstream tasks.
Then, recent works attempt to incorporate learnable operators to adapt clustering based on downstream tasks. For example, DiffPool \cite{yingHierarchicalGraphRepresentation2018} adopts GNN to learn an assignment matrix to assign nodes to clusters in the next layer for graph classification purposes. MinCutPool \cite{bianchiSpectralClusteringGraph2020} introduces a differentiable formulation of spectral clustering with multi-perceptron layers.
In this study, we utilize a similar approach to DiffPool by using GCN for the clustering process. But differently, we additionally consider the graph upsampling process to achieve node-level regression for traffic forecasting.



\section{Problem Definition}
Traffic forecasting can be seen as a time series forecasting problem with $N$ correlated time series data. At time step $t$, there are $\mathbf{X}_t=\{ x_t^1, x_t^2,\ldots, x_t^N \}^T \in \mathbb{R}^{N \times 1}$ recordings from $N$ sources of traffic data. Compared to the single-step traffic forecasting, where previous $T$ steps of data are used to predict the next step $\mathbf{X}_{t+1}$, multi-step traffic forecasting can provide more insights into how traffic conditions will shift from short-term to long-term. To perform multi-step traffic forecasting, the past $T$ steps and a function $\mathcal{F}$ given all the learnable parameters $\theta$ are used to predict the next $H$ steps:
\begin{equation}
    \{\mathbf{X}_{t+1}, \cdots, {\mathbf{X}_{t+H}} \} = \mathcal{F}_\theta \{ \mathbf{X}_{t},  \cdots, {\mathbf{X}_{t-T+1}}\}.
\end{equation}


To capture the spatial correlation between the traffic series, a graph $\mathcal{G} = (\mathcal{V, E}, \mathbf{A})$ is used in recent graph-based works to describe the road network, where $\mathcal{V}$ is the set of nodes generating the time series and $\vert \mathcal{V} \vert{}=N$, $\mathcal{E}$ is a set of edges connecting the nodes, and $\mathbf{A}$ is the adjacency matrix that describes the connection between the nodes (e.g., represented by road network distance or node similarity). Thus, the problem is modified as: 
\begin{equation}
    \{\mathbf{X}_{t+1}, \cdots, {\mathbf{X}_{t+H}} \} = \mathcal{F}_\theta \{ \mathbf{X}_{t}, \cdots, {\mathbf{X}_{t-T+1}} ; \mathcal{G}\}.
\end{equation}


\begin{figure}[htbp]
    \centering
    \includegraphics[width=0.99\linewidth]{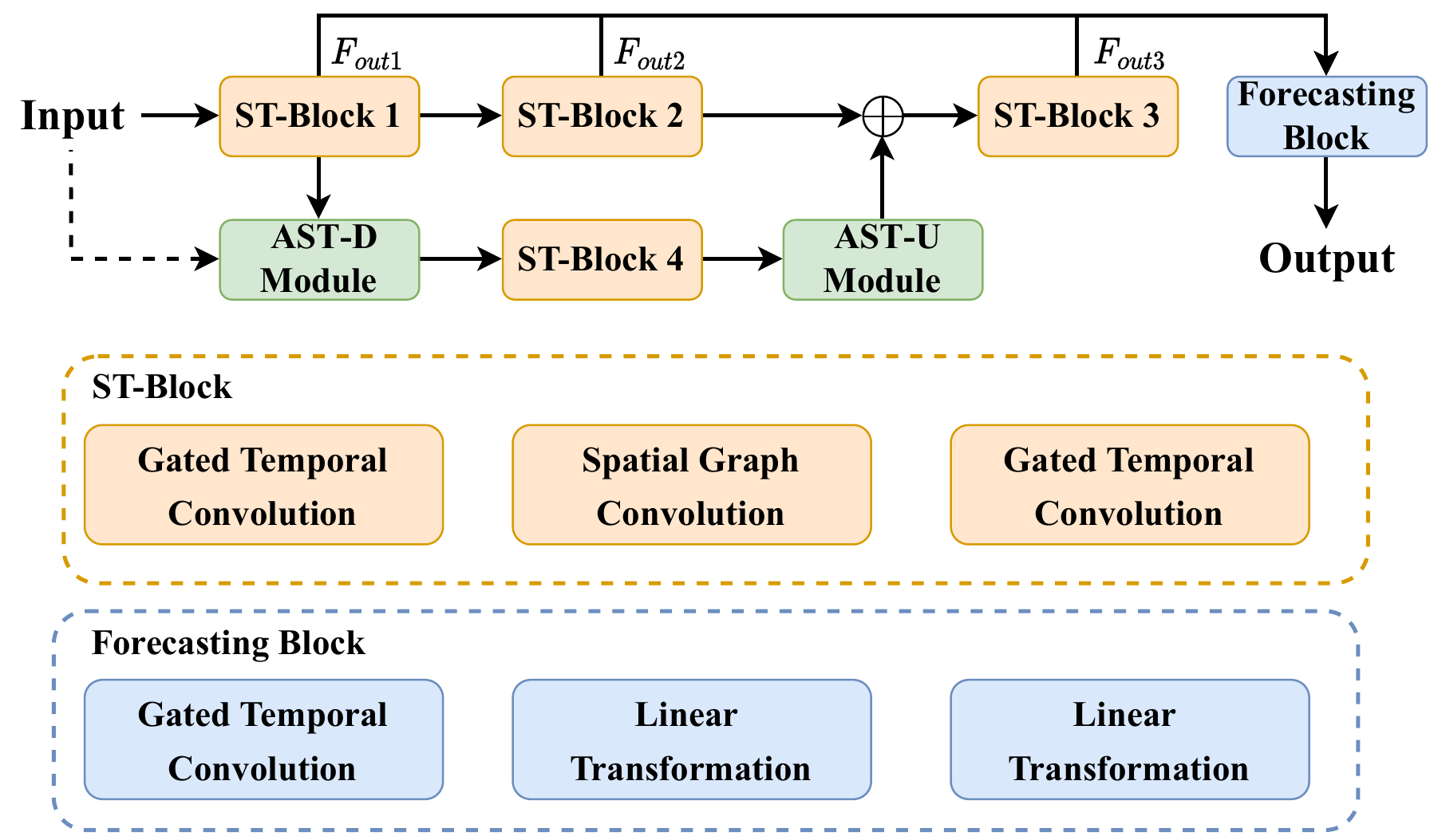}
    \caption{Illustration of our Adaptive Hierarchical Spatio-Temporal Network.}
    \label{fig:framework}
\end{figure}

\section{Methodology}
The overall framework of the proposed AHSTN is illustrated in Fig. \ref{fig:framework}. Similar to existing spatiotemporal models, AHSTN employs spatiotemporal blocks (detailed in Sec.\ref{sec:stblock}) to extract the spatial and temporal correlations in the traffic data. Different from previous works, AHSTN further models the spatial hierarchy and captures the spatiotemporal correlations at the cluster level with the help of the designed Adaptive SpatioTemporal Downsampling (AST-D) module and the Adaptive SpatioTemporal Upsampling (AST-U) module (elaborated in Sec.\ref{sec:hirer}). AHSTN merges the node-level and cluster-level together to capture the multi-scale spatiotemporal correlations for more accurate predictions. 

\subsection{SpatioTemporal Block}
\label{sec:stblock}
The SpatioTemporal Block (ST-Block) is employed to extract the spatial and temporal correlations from traffic data. It consists of three main parts: Gated Temporal Convolution Network (GTCN), GCN, and another GTCN, forming a sandwich structure as proposed in STGCN \cite{yuSpatioTemporalGraphConvolutional2018}. 
The ReLU activation function is employed after the GCN, followed by a batch normalization layer to standardize outputs across mini-batches.

Specifically, GTCN uses 1D-CNN on the time dimension with a filter size of $K$ to capture temporal correlation followed by a Gated Linear Unit (GLU) for non-linearity. Given the input $X \in \mathbb{R}^{N\times T \times C}$, gated temporal convolution is defined as:
\begin{align}
 \textit{TC}(X) ={}& \textit{Conv}(X) \in \mathbb{R}^{N\times (T-K+1) \times (C_t * 2)} \\
 (P, Q) ={}& \textit{Split}(\textit{TC}(X)), P,Q \in \mathbb{R}^{N\times (T-K+1) \times C_t} \\
 Z ={}& P \otimes \sigma (Q) \in \mathbb{R}^{N\times (T-K+1) \times C_t},
\end{align}
where $Z$ is the output of the GTCN,  $C_t$ is the channel size of the learned representation, $\sigma$ is the sigmoid activation function, and $\otimes$ is the element-wise Hadamard product. 

Graph convolution follows the 1st-order approximation of Chebyshev spectral filter proposed in \cite{kipfSemiSupervisedClassificationGraph2017} to aggregate information from neighboring nodes. Given the input, $X \in \mathbb{R}^{N\times T \times C}$, the output $Z$ of the graph convolution capturing the node interactions is defined as:
\begin{equation}
    Z = \sigma (\Tilde{D}^{-\frac{1}{2}} \Tilde{A}\Tilde{D}^{-\frac{1}{2}} X W) \in \mathbb{R}^{N\times T \times C_g},
\end{equation}
where $Z$ is the learned node embedding, $\Tilde{A} = A + I$ to account for self-loop, $\Tilde{D} = \sum_j \Tilde{A}_{ij}$ to normalize the $\Tilde{A}$, $W$ is the trainable parameter, and $C_g$ is the channel size for graph convolution output.

\subsection{Spatiotemporal Correlations in the Cluster-level} \label{sec:hirer}
In this section, we introduce how to model the spatial hierarchy of the road network and learn spatiotemporal correlations at the cluster level. 
Specifically, we first introduce the AST-D module as the foundation for modeling the spatial hierarchy, which adaptively learns an assignment matrix to merge the nodes (e.g., road segments or intersections) in the road network into high-level clusters (e.g., similar semantic regions). Based on the graph clustering results, cluster-level spatiotemporal correlations can be extracted by spatiotemporal blocks based on the cluster-level graph and representations. Next, an AST-U module is utilized to project the cluster-level spatiotemporal representation back to the node level.

\textbf{AST-D module}: 
The AST-D module achieves the graph downsampling operation considering the spatiotemporal correlations in the traffic data and generates the cluster-level graph and representations. 
As shown in Fig. \ref{fig:framework}, AST-D utilizes a separate GCN (i.e., \textit{clustering GCN}) to learn a mapping matrix $M \in \mathbb{R}^{N \times N'}$ between $N$ nodes and $N'$ clusters based on node features $X \in \mathbb{R}^{N\times T \times C}$ and node-level adjacency matrix $A \in \mathbb{R}^{N\times N}$. Instead of using the hard matching method, $M$ is a \textbf{soft} assignment matrix that indicates the probability of a node belonging to a cluster.
The number of clusters $N'$ is adjusted by the clustering ratio $P_\text{cluster}$, which is a hyper-parameter and related to the road network size and characteristics. Mathematically,
 \begin{align}
 N' &= N \times P_\text{cluster} \\
 M &= \textit{GCN}_{\text{cluster}}(X, A) \in \mathbb{R}^{N\times N'}\\
 M &= \textit{Softmax}(\frac{M}{\tau}),
 \end{align}
where the $\textit{Softmax}(\cdot)$ function is applied row-wise to ensure the probability of a node mapping to all clusters is 1, and $\tau$ is a temperature parameter used to adjust the output distribution of the $\textit{Softmax}(\cdot)$. While the assignment matrix could be dynamic with respect to different inputs and time, in this work, we consider the cluster-level connectivity to be static and thus conduct a mean aggregation to $M$ on the batch axis before the $\textit{Softmax}(\cdot)$. Moreover, the exponential moving average can be applied to $M$ during the whole training period to further stabilize the training process.

The assignment matrix $M$ is then applied to the node-level feature embedding learned from ST-Block $Z_{\text{node}} \in \mathbb{R}^{N\times T \times C}$ and the original node-level adjacency matrix $A$ to generate the corresponding cluster-level embedding $Z_{\text{cluster}} $ and adjacency matrix $A_\text{cluster} \in \mathbb{R}^{N' \times N'}$ as follows:
\begin{align}
    Z_{\text{cluster}} ={}& M^T \cdot Z_{\text{node}} \in \mathbb{R}^{N'\times T \times C} \\
    A_\text{cluster}  ={}& M^T \cdot A \cdot M \in \mathbb{R}^{N' \times N'}.
\end{align}

\textbf{Cluster-level Spatiotemporal Learning} Based on $Z_{\text{cluster}}$ and $A_\text{cluster}$, spatiotemporal correlations can be captured in the cluster-level via another ST-Block. The detailed calculation of the cluster-level ST-Block is the same as the ST-Block described in Section.\ref{sec:stblock}, except for the number of nodes in $Z_{\text{cluster}}$ and $A_\text{cluster}$ is $N'$.

\textbf{AST-U module}: After learning the spatiotemporal correlations in the cluster level with the ST-Block, it is necessary to project the learned cluster-level representation back to the node-level representation for prediction via upsampling. To achieve this goal, an AST-U module is introduced as follows:
\begin{align}
Z_{\text{node}} ={}& M^+ \cdot Z_{\text{cluster}} \in \mathbb{R}^{N\times T \times C}
\end{align}
where $M^+ \in \mathbb{R}^{N' \times N}$ is the inverse of the assignment matrix $M$. In implementation, the Moore–Penrose inverse is used to get $M^+$ from $M$ considering its differentiable nature.

\subsection{Momentum Update}      
To stabilize the learning of the clustering matrix $M$, the momentum update is desired. Due to the limit of mini-batch learning, the clustering GCN can only see a small set of data at each iteration. With the fluctuation and noise in traffic conditions, each mini-batch $\mathbf{X}$ is unlikely to represent the whole picture of how nodes should be clustered together. Therefore, the clustering matrix $M \in \mathbb{R}^{N \times N'}$ is first randomly initialized. At each iteration, we update $M$ with clustering matrix $M' \in \mathbb{R}^{N \times N'}$ learned from the mini-batch $\mathbf{X}$ by:
\begin{equation}
    M \leftarrow \alpha M + (1-\alpha) M'.
\end{equation}
Here, $\alpha$ is the momentum update hyperparameter that controls how much information from $M'$ is allowed into $M$. Only $M'$ is back-propagated during training. Once training is finished, the clustering matrix $M$ is saved within the model and not updated during inference.

\subsection{Traffic Forecasting and Optimization}
The traffic forecasting block takes feature embeddings learned from both the node and cluster levels to produce the final traffic forecast. Additionally, to capture features at different time scales, temporal skip connections are used to concatenate outputs from different ST-Blocks ($F_{\text{out}1}$, $F_{\text{out}2}$, $F_{\text{out}3}$ in Fig.\ref{fig:framework}) on the feature dimension, resulting in the final spatiotemporal representation capturing both multi-scale spatial correlations and multi-scale temporal correlations. As shown in Fig. \ref{fig:framework}, the final representation for forecasting is:
\begin{align}
    F_\text{final} = \textit{Concat}(F_{\text{out}1}[:,:,-T_3 :], 
    F_{\text{out}2}[:,:,-T_3 :], F_{\text{out}3})\\
      {{}} \in \mathbb{R}^{B \times N \times T_3 \times (C_t \times 3)} \notag
\end{align}
where $T_3$ is the temporal channel dimension after a set of temporal convolutions in the spatiotemporal blocks. 

Finally, $F_{final}$ is fed into the forecasting block, comprised of a gated temporal convolution layer and two layers of linear transformation with ReLU non-linearity, to generate prediction $\hat{X}_{t+1:t+H}$ in the following $H$ steps. 
During training, the Mean Absolute Error between the prediction $\hat{\mathbf{X}}_{t+1:t+H}$ and the groundtruth $\mathbf{X}_{t+1:t+H}$ is used as loss function for the network optimzation:
\begin{equation}
    L(W_\theta) = \sum^{i=t+H}_{i=t+1}|\mathbf{X}_i - \hat{\mathbf{X}}_i|,
\end{equation}
where $W_\theta$ is all the trainable parameters.

\section{Experiments}
In this section, the performance of the proposed AHSTN is evaluated on two real-world traffic datasets. We also carried out ablation studies to showcase the effectiveness of the proposed mechanism in AHSTN.

\textbf{Datasets}
Two real-world traffic datasets released in \cite{guoHierarchicalGraphConvolution2021} are collected by Didi Chuxing GAIA Initiative from Jinan, China and Xian, China. The datasets contain the average speed of road segments in Jinan and Xian, which are aggregated by a 10-minute time span. Both datasets have 52,286 samples, which cover about one year in length. There are 561 and 792 nodes (road segments) respectively in Jinan and Xian datasets. The adjacency matrices for both Jinan and Xian are constructed using a threshold Gaussian kernel based on the distance between the road segments \cite{guoHierarchicalGraphConvolution2021}.

\textbf{Comparison Methods and Evaluation Metrics }
The proposed AHSTN is compared with the following methods: \begin{enumerate*} \item Historical Average (HA), \item ARIMA, \item LSTM \cite{kangShorttermTrafficFlow2017}, \item STGCN \cite{yuSpatioTemporalGraphConvolutional2018}, \item GWNET \cite{wuGraphWaveNetDeep2019}, \item AGCRN \cite{baiAdaptiveGraphConvolutional2020}, \item HGCN \cite{guoHierarchicalGraphConvolution2021}, \item STG-NCDE \cite{choiGraphNeuralControlled2022}, \item TFormer \cite{yanLearningDynamicHierarchical2022} \end{enumerate*}. To compare the forecasting performance, the widely used metrics Mean Absolute Error (MAE), Root Mean Squared Error (RMSE), and Mean Absolute Percentage Error (MAPE) are adopted. Time steps with missing values are excluded from the evaluation. 

\textbf{Training Details}
All experiments are implemented and tested using Pytorch 1.10.0 on a workstation with Nvidia RTX2080Ti. We used the previous 12 time steps' data to forecast traffic conditions in the next 12 time steps. 
The Z-score normalization is used to pre-process the datasets to facilitate the training
The batch size used is 64. The $C_t$ temporal and $C_g$ graph channel size of ST-Blocks used is 64 and 32, respectively. The clustering ratio $P_\text{cluster}$ is 0.1 for Jinan and 0.06 for Xian. Thus, there are 56 clusters for Jinan and 48 clusters for Xian. We train AHSTN using the Adam optimizer for 100 epochs. The learning rate is set to 0.002 at the start with exponential decay of 0.99. 
To promote the generalization ability of AHSTN, we further train the model based on the best validation model on both training and validation datasets for three epochs. 

\begin{table*}[htbp]
\centering
\caption{Performance comparison of AHSTN and other models on Jinan and Xian dataset. The best results are highlighted in bold.}
\label{tab:result}
\begin{tabular}{ccccccccccc}
\hline
\multirow{2}{*}{Dataset} & \multirow{2}{*}{Method} & \multicolumn{3}{c}{30 min} & \multicolumn{3}{c}{1 hour} & \multicolumn{3}{c}{2 hour} \\ \cline{3-11} 
 &  & MAE & MAPE & RMSE & MAE & MAPE & RMSE & MAE & MAPE & RMSE \\ \hline
\multirow{10}{*}{Jinan} & HA & 5.69 & 20.02\% & 7.60 & 5.69 & 20.02\% & 7.60 & 5.69 & 20.02\% & 7.60 \\
 & ARIMA & 3.96 & 14.12\% & 6.14 & 4.48 & 16.10\% & 6.56 & 5.09 & 18.24\% & 7.01 \\
 & LSTM & 3.21 & 12.85\% & 4.85 & 3.67 & 14.92\% & 5.48 & 4.30 & 17.38\% & 6.26 \\
 & STGCN & 2.93 & 11.82\% & 4.46 & 3.26 & 13.22\% & 4.89 & 3.65 & 14.80\% & 5.37 \\
 & GWNET & 2.89 & 11.59\% & 4.37 & 3.13 & 12.50\% & 4.65 & 3.50 & 13.70\% & 5.11 \\
 & AGCRN & 2.90 & 11.49\% & 4.39 & 3.15 & 12.48\% & 4.70 & 3.50 & 13.74\% & 5.13 \\
 & HGCN & 2.90 & 11.56\% & 4.40 & 3.11 & 12.45\% & 4.70 & 3.36 & 13.40\% & 5.02 \\
 & \multicolumn{1}{l}{STG-NCDE} & 3.45 & 12.89\% & 5.94 & 3.70 & 14.20\% & 6.14 & 3.93 & 15.07\% & 6.21 \\
 & TFormer & 3.10 & 12.26\% & 4.66 & 3.39 & 13.42\% & 5.00 & 3.92 & 15.30\% & 5.58 \\
 & AHSTN & \textbf{2.78} & \textbf{11.15\%} & \textbf{4.21} & \textbf{2.98} & \textbf{12.16\%} & \textbf{4.54} & \textbf{3.28} & \textbf{13.03\%} & \textbf{4.88} \\ \hline
\multirow{10}{*}{Xian} & HA & 6.02 & 21.79\% & 8.16 & 6.02 & 21.79\% & 8.16 & 6.02 & 21.79\% & 8.16 \\
 & ARIMA & 3.70 & 12.96\% & 6.05 & 4.26 & 15.28\% & 6.57 & 5.04 & 18.26\% & 7.24 \\
 & LSTM & 3.16 & 11.92\% & 4.83 & 3.70 & 14.22\% & 5.52 & 4.52 & 17.42\% & 6.53 \\
 & STGCN & 2.85 & 10.79\% & 4.40 & 3.19 & 12.33\% & 4.84 & 3.65 & 14.07\% & 5.42 \\
 & GWNET & 2.76 & 10.46\% & 4.26 & 3.03 & 11.71\% & 4.61 & 3.44 & 13.22\% & 5.10 \\
 & AGCRN & 2.77 & 10.47\% & 4.26 & 3.01 & 11.65\% & 4.58 & 3.38 & 13.07\% & 5.04 \\
 & HGCN & 2.75 & 10.43\% & 4.27 & 2.97 & 11.50\% & 4.57 & 3.23 & 12.57\% & 4.88 \\
 & STG-NCDE & 3.34 & 12.13\% & 5.84 & 3.62 & 13.50\% & 6.11 & 4.07 & 15.28\% & 6.60 \\
 & TFormer & 2.89 & 11.10\% & 4.42 & 3.18 & 12.39\% & 4.80 & 3.60 & 14.25\% & 5.38 \\
 & AHSTN & \textbf{2.71} & \textbf{10.24\%} & \textbf{4.24} & \textbf{2.92} & \textbf{11.34\%} & \textbf{4.52} & \textbf{3.20} & \textbf{12.52\%} & \textbf{4.87} \\ \hline
\end{tabular}
\end{table*}

\subsection{Experiment Result}
Table \ref{tab:result} illustrates the performance of AHSTN in comparison to other models for 30-minute, one-hour, and two-hour traffic forecasting on Jinan and Xian datasets. It can be observed that both the proposed AHSTN and HGCN achieve better performance than non-GCN models and competitive GCN models such as STGCN, Graph Wavenet, and AGCRN, which use only node-level information from the road network graph. STG-NCDE shows poor performance in both datasets due to its preprocessing step involving fitting a natural cubic spline to the traffic data. The high fluctuations present in urban traffic speed can cause the cubic spline to overfit the data leading to poor generalizability. Transformer-based TFormer's performance is hindered by its local-global interaction using only the K-hop adjacent mask, which focuses on the node level alone. The results demonstrate the effectiveness of modeling the spatial hierarchy in traffic data. When compared with HGCN which uses a pre-learned spatial hierarchy, AHSTN shows a clear advantage for short- and medium-term prediction (30 minutes and 60 minutes) owning to the multi-scale spatiotemporal representations obtained with the adaptive hierarchical learning. For long-term predictions (two hours), AHSTN shows a comparable or superior result to HGCN.
Arguably, short-term traffic forecasting accuracy is more important to ITS as many ITS functions such as route planning and traffic optimization rely on more accurate short-term traffic forecasting results.

\textbf{Computational Cost} To evaluate the computation cost, a comparison of computational time for both training (one epoch) and inference among STGCN, GWNET, HGCN, TFormer, and AHSTN is presented in Table.\ref{tab:comp_time}. We also included the number of parameters in each model for comparison. STGCN has the lowest training and inference time. However, STGCN does not model the spatial hierarchy which limits its forecasting performance. While HGCN considers the spatial hierarchy and achieves great performance, its' training cost is about two times of STGCN. Additionally, HGCN uses pre-learned clustering information which further complicates the training process. TFormer's encoder-decoder architecture greatly increases its training and inference time with a limited performance increase.
In comparison, the proposed AHSTN learns the clustering on-the-fly and is about 25\% faster in training compared to HGCN while achieving superior performance on both datasets. In summary, considering both the computation time and the performance presented in Table. \ref{tab:result}, AHSTN achieves excellent performance with balanced computation cost.

\begin{table}[htbp]
\centering
\caption{Computational cost comparison of AHSTN and other GCN models.}
\label{tab:comp_time}
\begin{tabular}{ccccc}
\hline
\multirow{2}{*}{Dataset} & \multirow{2}{*}{Method} & \multicolumn{2}{c}{Computational Time (s)} & \multirow{2}{*}{\# Parameters} \\ \cline{3-4}
 &  & Training & Inference &  \\ \hline
\multirow{5}{*}{Jinan} & STGCN & 33 & 4 & 424,833 \\
 & GWNET & 145 & 15 & 316,448 \\
 & HGCN & 63 & 6 & 1,021,728 \\
 & TFormer & 432 & 36 & 852,172 \\
 & AHSTN & 46 & 5 & 305,370 \\ \hline
\multirow{5}{*}{Xian} & STGCN & 47 & 5 & 543,105 \\
 & GWNET & 184 & 20 & 321,068 \\
 & HGCN & 92 & 8 & 1,418,437 \\
 & TFormer & 502 & 42 & 852,172 \\
 & AHSTN & 70 & 7 & 304,850 \\ \hline
\end{tabular}
\end{table}

\begin{figure}[htbp]
\begin{minipage}[c]{0.45\linewidth}
\centering
\includegraphics[width=\textwidth]{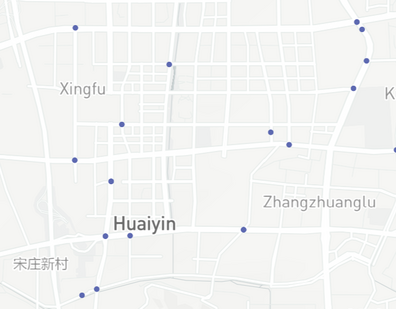}
(a)\label{fig:node-cluster}
\end{minipage}
\begin{minipage}[c]{0.45\linewidth}
\centering
\includegraphics[width=\textwidth,height=3cm]{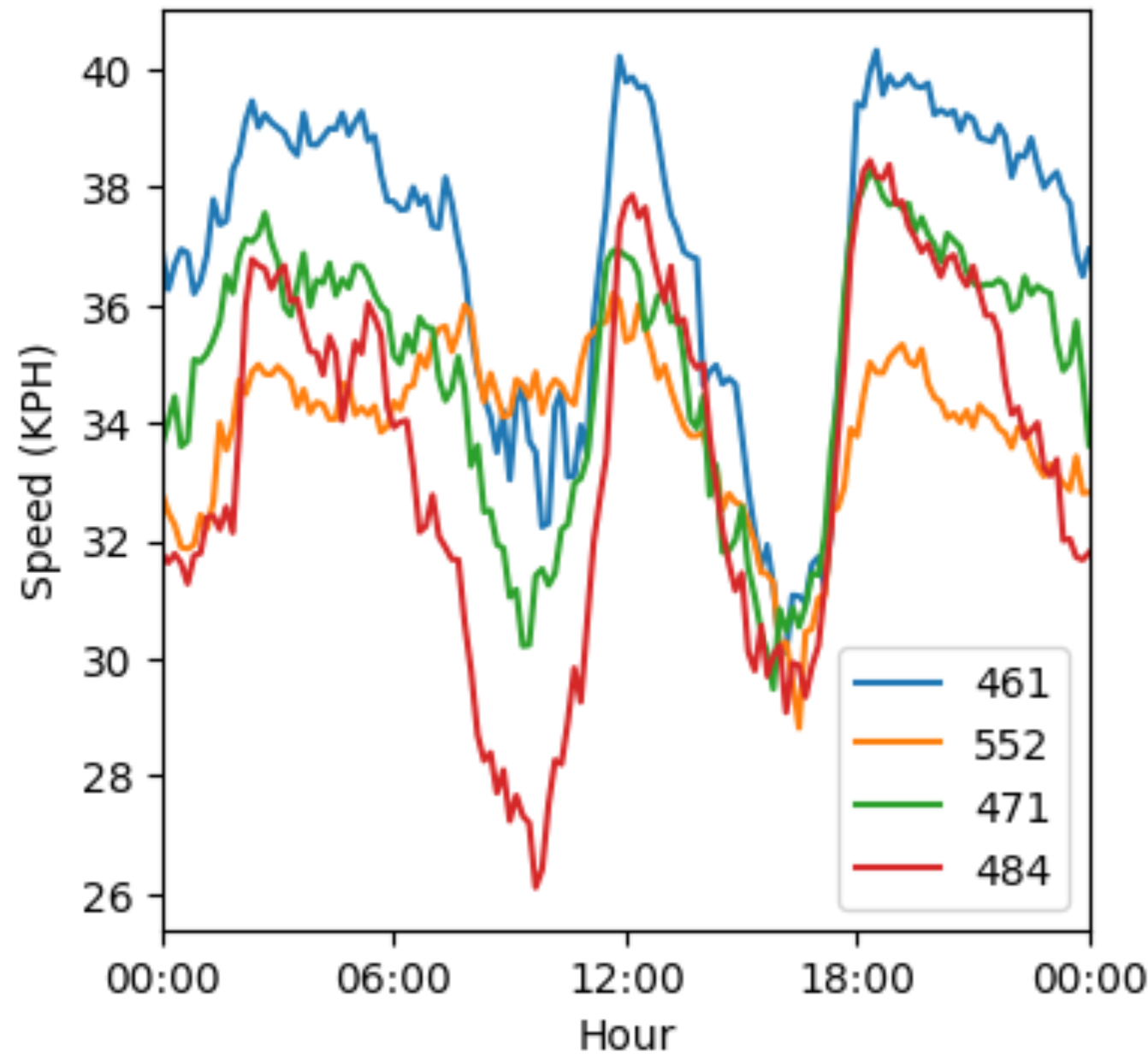}
(b)\label{fig:node-speed}
\end{minipage}
\caption{Clustering Visualization on Jinan Dataset.} \label{fig:cluster}
\end{figure}

\begin{table*}[t]
\centering
\caption{Performance comparison of AHSTN and alternated architectures for ablation study.}
\label{tab:ablation}
\begin{tabular}{ccccccccccc}
\hline
\multirow{2}{*}{Dataset} & \multirow{2}{*}{Method} & \multicolumn{3}{c}{30 min} & \multicolumn{3}{c}{1 hour} & \multicolumn{3}{c}{2 hour} \\ \cline{3-11} 
 &  & MAE & MAPE & RMSE & MAE & MAPE & RMSE & MAE & MAPE & RMSE \\ \hline
Jinan & AHSTN & \textbf{2.78} & \textbf{11.15\%} & \textbf{4.21} & \textbf{2.98} & \textbf{12.16\%} & \textbf{4.54} & \textbf{3.28} & \textbf{13.03\%} & \textbf{4.88} \\
 & AHSTN-S & 2.81 & 11.25\% & 4.28 & 3.04 & 12.23\% & 4.61 & 3.34 & 13.43\% & 4.98 \\
 & AHSTN-H & 2.81 & 11.30\% & 4.30 & 3.06 & 12.34\% & 4.62 & 3.37 & 12.58\% & 5.01 \\ \hline
Xian & AHSTN & \textbf{2.71} & \textbf{10.24\%} & \textbf{4.24} & \textbf{2.92} & \textbf{11.34\%} & \textbf{4.52} & \textbf{3.20} & \textbf{12.52\%} & \textbf{4.87} \\
 & AHSTN-S & 2.74 & 10.44\% & 4.26 & 2.99 & 11.62\% & 4.59 & 3.31 & 12.94\% & 5.01 \\
 & AHSTN-H & 2.73 & 10.46\% & 4.26 & 2.99 & 11.75\% & 4.62 & 3.34 & 13.20\% & 5.08 \\ \hline
\end{tabular}
\end{table*}

\textbf{Adaptive Clustering Visualization} Fig. \ref{fig:cluster} visualizes the effect of the proposed AST-D module on the Jinan dataset. The intersections shown in Fig. \ref{fig:cluster}(a) are soft clustered together to form a region, after which the spatiotemporal correlations are extracted together with ST-Block4. Fig. \ref{fig:cluster}(b) shows the average traffic speed over 24 hours from four randomly chosen intersections within this formed cluster. We can observe that they share similar trends and variations in traffic conditions. 

\subsection{Ablation Study}

To further examine the effectiveness of different modules in AHSTN, ablation studies are conducted by removing key modules. Specifically, we construct, 1) AHSTN-S by removing all temporal skip connections after each ST-Block, i.e., directly using $F_{out3}$ for prediction, 2) AHSTN-H by removing the spatial hierarchy learning parts, i.e., removing AST-D, ST-Block4, and AST-U.
The performance comparison between these alternated architectures and AHSTN is shown in Table.\ref{tab:ablation}. As can be observed, AHSTN achieves the best performance among all variants, which demonstrates the necessity of learning multi-scale spatial and temporal correlations. 

\subsection{Hype-parameter Analysis}

\begin{figure}[htbp]
    \centering
    \includegraphics[width=1.0\linewidth]{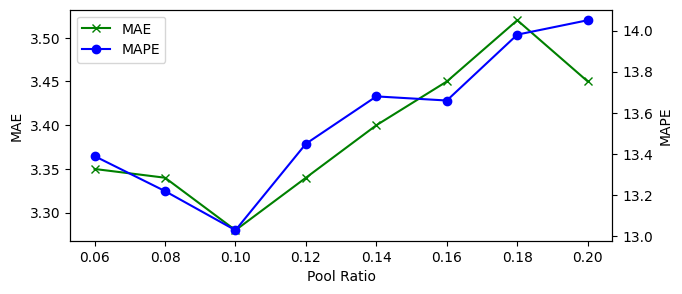}
    \caption{MAE and MAPE of 2-hour forecasting result using AHSTN with different clustering ratio $P_\text{cluster}$ on Jinan dataset}
    \label{fig:hparm}
\end{figure}

In this section, the effect of clustering ratio $P_\text{cluster}$ is examined. As a hyper-parameter, $P_\text{cluster}$ determines how many clusters are formed with the adaptive spatiotemporal downsampling module from the original graph. As can be seen from Fig. \ref{fig:hparm}, balancing the clustering ratio $P_\text{cluster}$ is required to achieve the best forecasting result. For the Jinan dataset, a clustering ratio of 0.1 helps achieve the best performance, which pools 561 original nodes to 56 clusters. 
A smaller clustering ratio can lead to extremely dense clusters where many nodes are pooled into one cluster, thus losing out on the information carried by individual nodes. In contrast, a bigger clustering ratio leads to over-coarse clusters, where many clusters only contain one node, and thus cannot capture appropriate spatial hierarchies.

\section{Conclusion}
In this paper, a novel and end-to-end spatial hierarchy learning GCN framework, AHSTN, is proposed for the traffic forecasting problem. 
The proposed method introduces adaptive spatiotemporal downsampling for learning cluster-level spatiotemporal representations and adaptive spatiotemporal upsampling module for obtaining multi-scale spatiotemporal correlations. Experiments on two real-world datasets show AHSTN outperforms related state-of-the-art traffic forecasting methods for both short-term and long-term traffic forecasting. AHSTN also achieves faster training and uses fewer parameters than the previous study using spatial hierarchy. In future works, the adaptive hierarchy learning mechanism can be further explored to form superclusters by downsampling on the formed clusters to explore more complicated spatial hierarchies in the urban road networks.

\addtolength{\textheight}{-10cm} 

\bibliographystyle{IEEEtran}
\bibliography{main}

\begin{thebibliography}{10}
\providecommand{\url}[1]{#1}
\csname url@samestyle\endcsname
\providecommand{\newblock}{\relax}
\providecommand{\bibinfo}[2]{#2}
\providecommand{\BIBentrySTDinterwordspacing}{\spaceskip=0pt\relax}
\providecommand{\BIBentryALTinterwordstretchfactor}{4}
\providecommand{\BIBentryALTinterwordspacing}{\spaceskip=\fontdimen2\font plus
\BIBentryALTinterwordstretchfactor\fontdimen3\font minus
  \fontdimen4\font\relax}
\providecommand{\BIBforeignlanguage}[2]{{%
\expandafter\ifx\csname l@#1\endcsname\relax
\typeout{** WARNING: IEEEtran.bst: No hyphenation pattern has been}%
\typeout{** loaded for the language `#1'. Using the pattern for}%
\typeout{** the default language instead.}%
\else
\language=\csname l@#1\endcsname
\fi
#2}}
\providecommand{\BIBdecl}{\relax}
\BIBdecl

\bibitem{wang2023correlated}
L.~Wang, L.~Bai, Z.~Li, R.~Zhao, and F.~Tsung, ``Correlated time series
  self-supervised representation learning via spatiotemporal bootstrapping,''
  \emph{arXiv preprint arXiv:2306.06994}, 2023.

\bibitem{li2022profile}
Z.~Li, H.~Yan, F.~Tsung, and K.~Zhang, ``Profile decomposition based hybrid
  transfer learning for cold-start data anomaly detection,'' \emph{ACM
  Transactions on Knowledge Discovery from Data (TKDD)}, vol.~16, no.~6, pp.
  1--28, 2022.

\bibitem{lan2023mm}
T.~Lan, Z.~Li, Z.~Li, L.~Bai, M.~Li, F.~Tsung, W.~Ketter, R.~Zhao, and
  C.~Zhang, ``Mm-dag: Multi-task dag learning for multi-modal data--with
  application for traffic congestion analysis,'' \emph{arXiv preprint
  arXiv:2306.02831}, 2023.

\bibitem{li2020tensor}
Z.~Li, N.~D. Sergin, H.~Yan, C.~Zhang, and F.~Tsung, ``Tensor completion for
  weakly-dependent data on graph for metro passenger flow prediction,'' in
  \emph{proceedings of the AAAI conference on artificial intelligence},
  vol.~34, no.~04, 2020, pp. 4804--4810.

\bibitem{li2020long}
Z.~Li, H.~Yan, C.~Zhang, and F.~Tsung, ``Long-short term spatiotemporal tensor
  prediction for passenger flow profile,'' \emph{IEEE Robotics and Automation
  Letters}, vol.~5, no.~4, pp. 5010--5017, 2020.

\bibitem{kamarianakisForecastingTrafficFlow2003}
Y.~Kamarianakis and P.~Prastacos, ``Forecasting {{Traffic Flow Conditions}} in
  an {{Urban Network}}: {{Comparison}} of {{Multivariate}} and {{Univariate
  Approaches}},'' \emph{Transportation Research Record}, vol. 1857, no.~1, pp.
  74--84, Jan. 2003.

\bibitem{williamsModelingForecastingVehicular2003}
B.~M. Williams and L.~A. Hoel, ``Modeling and {{Forecasting Vehicular Traffic
  Flow}} as a {{Seasonal ARIMA Process}}: {{Theoretical Basis}} and {{Empirical
  Results}},'' \emph{Journal of Transportation Engineering}, vol. 129, no.~6,
  pp. 664--672, Nov. 2003.

\bibitem{zhangCombiningWeatherCondition2017}
D.~Zhang and M.~R. Kabuka, ``Combining {{Weather Condition Data}} to {{Predict
  Traffic Flow}}: {{A GRU Based Deep Learning Approach}},'' in \emph{2017
  {{IEEE}} 15th {{Intl Conf}} on {{Dependable}}, {{Autonomic}} and {{Secure
  Computing}}, 15th {{Intl Conf}} on {{Pervasive Intelligence}} and
  {{Computing}}, 3rd {{Intl Conf}} on {{Big Data Intelligence}} and
  {{Computing}} and {{Cyber Science}} and {{Technology
  Congress}}({{DASC}}/{{PiCom}}/{{DataCom}}/{{CyberSciTech}})}, Nov. 2017, pp.
  1216--1219.

\bibitem{yuDeepLearningGeneric2017}
R.~Yu, Y.~Li, C.~Shahabi, U.~Demiryurek, and Y.~Liu, ``Deep {{Learning}}: {{A
  Generic Approach}} for {{Extreme Condition Traffic Forecasting}},'' in
  \emph{Proceedings of the 2017 {{SIAM International Conference}} on {{Data
  Mining}} ({{SDM}})}, ser. Proceedings.\hskip 1em plus 0.5em minus 0.4em\relax
  {Society for Industrial and Applied Mathematics}, Jun. 2017, pp. 777--785.

\bibitem{kangShorttermTrafficFlow2017}
D.~Kang, Y.~Lv, and Y.-y. Chen, ``Short-term traffic flow prediction with
  {{LSTM}} recurrent neural network,'' in \emph{2017 {{IEEE}} 20th
  {{International Conference}} on {{Intelligent Transportation Systems}}
  ({{ITSC}})}, Oct. 2017, pp. 1--6.

\bibitem{yuSpatioTemporalGraphConvolutional2018}
B.~Yu, H.~Yin, and Z.~Zhu, ``Spatio-{{Temporal Graph Convolutional Networks}}:
  {{A Deep Learning Framework}} for {{Traffic Forecasting}},'' in
  \emph{Proceedings of the {{Twenty-Seventh International Joint Conference}} on
  {{Artificial Intelligence}}}.\hskip 1em plus 0.5em minus 0.4em\relax
  {Stockholm, Sweden}: {International Joint Conferences on Artificial
  Intelligence Organization}, Jul. 2018, pp. 3634--3640.

\bibitem{zhangDeepSpatioTemporalResidual2017}
J.~Zhang, Y.~Zheng, and D.~Qi, ``Deep {{Spatio-Temporal Residual Networks}} for
  {{Citywide Crowd Flows Prediction}},'' \emph{Proceedings of the AAAI
  Conference on Artificial Intelligence}, vol.~31, no.~1, Feb. 2017.

\bibitem{jiangGeospatialDataImages2019}
W.~Jiang and L.~Zhang, ``Geospatial data to images: {{A}} deep-learning
  framework for traffic forecasting,'' \emph{Tsinghua Science and Technology},
  vol.~24, no.~1, pp. 52--64, Feb. 2019.

\bibitem{geng2019spatiotemporal}
X.~Geng, Y.~Li, L.~Wang, L.~Zhang, Q.~Yang, J.~Ye, and Y.~Liu, ``Spatiotemporal
  multi-graph convolution network for ride-hailing demand forecasting,'' in
  \emph{Proceedings of the AAAI conference on artificial intelligence},
  vol.~33, no.~01, 2019, pp. 3656--3663.

\bibitem{ziyue2021tensor}
L.~Ziyue, ``Tensor topic models with graphs and applications on individualized
  travel patterns,'' in \emph{2021 IEEE 37th International Conference on Data
  Engineering (ICDE)}.\hskip 1em plus 0.5em minus 0.4em\relax IEEE, 2021, pp.
  2756--2761.

\bibitem{li2022individualized}
Z.~Li, H.~Yan, C.~Zhang, and F.~Tsung, ``Individualized passenger travel
  pattern multi-clustering based on graph regularized tensor latent dirichlet
  allocation,'' \emph{Data Mining and Knowledge Discovery}, vol.~36, no.~4, pp.
  1247--1278, 2022.

\bibitem{liDiffusionConvolutionalRecurrent2018}
Y.~Li, R.~Yu, C.~Shahabi, and Y.~Liu, ``Diffusion {{Convolutional Recurrent
  Neural Network}}: {{Data-Driven Traffic Forecasting}},'' in
  \emph{International {{Conference}} on {{Learning Representations}}}, 2018.

\bibitem{wuGraphWaveNetDeep2019}
Z.~Wu, S.~Pan, G.~Long, J.~Jiang, and C.~Zhang, ``Graph {{WaveNet}} for {{Deep
  Spatial-Temporal Graph Modeling}},'' in \emph{Proceedings of the
  {{Twenty-Eighth International Joint Conference}} on {{Artificial
  Intelligence}}, {{IJCAI-19}}}.\hskip 1em plus 0.5em minus 0.4em\relax
  {International Joint Conferences on Artificial Intelligence Organization},
  Jul. 2019, pp. 1907--1913.

\bibitem{baiAdaptiveGraphConvolutional2020}
L.~Bai, L.~Yao, C.~Li, X.~Wang, and C.~Wang, ``Adaptive {{Graph Convolutional
  Recurrent Network}} for {{Traffic Forecasting}},'' in \emph{Advances in
  {{Neural Information Processing Systems}}}, vol.~33.\hskip 1em plus 0.5em
  minus 0.4em\relax {Curran Associates, Inc.}, 2020, pp. 17\,804--17\,815.

\bibitem{lin2023dynamic}
J.~Lin, Z.~Li, Z.~Li, L.~Bai, R.~Zhao, and C.~Zhang, ``Dynamic causal graph
  convolutional network for traffic prediction,'' \emph{arXiv preprint
  arXiv:2306.07019}, 2023.

\bibitem{guoHierarchicalGraphConvolution2021}
K.~Guo, Y.~Hu, Y.~Sun, S.~Qian, J.~Gao, and B.~Yin, ``Hierarchical {{Graph
  Convolution Network}} for {{Traffic Forecasting}},'' \emph{Proceedings of the
  AAAI Conference on Artificial Intelligence}, vol.~35, no.~1, pp. 151--159,
  May 2021.

\bibitem{jiangGraphNeuralNetwork2022}
W.~Jiang and J.~Luo, ``Graph neural network for traffic forecasting: {{A}}
  survey,'' \emph{Expert Systems with Applications}, vol. 207, p. 117921, Nov.
  2022.

\bibitem{grattarolaUnderstandingPoolingGraph2022}
D.~Grattarola, D.~Zambon, F.~M. Bianchi, and C.~Alippi, ``Understanding
  {{Pooling}} in {{Graph Neural Networks}},'' \emph{IEEE Transactions on Neural
  Networks and Learning Systems}, pp. 1--11, 2022.

\bibitem{dhillonWeightedGraphCuts2007}
I.~S. Dhillon, Y.~Guan, and B.~Kulis, ``Weighted {{Graph Cuts}} without
  {{Eigenvectors A Multilevel Approach}},'' \emph{IEEE Transactions on Pattern
  Analysis and Machine Intelligence}, vol.~29, no.~11, pp. 1944--1957, Nov.
  2007.

\bibitem{bianchiHierarchicalRepresentationLearning2022}
F.~M. Bianchi, D.~Grattarola, L.~Livi, and C.~Alippi, ``Hierarchical
  {{Representation Learning}} in {{Graph Neural Networks With Node Decimation
  Pooling}},'' \emph{IEEE Transactions on Neural Networks and Learning
  Systems}, vol.~33, no.~5, pp. 2195--2207, May 2022.

\bibitem{maGraphConvolutionalNetworks2019}
Y.~Ma, S.~Wang, C.~C. Aggarwal, and J.~Tang, ``Graph {{Convolutional Networks}}
  with {{EigenPooling}},'' in \emph{Proceedings of the 25th {{ACM SIGKDD
  International Conference}} on {{Knowledge Discovery}} \& {{Data Mining}}},
  ser. {{KDD}} '19.\hskip 1em plus 0.5em minus 0.4em\relax {New York, NY, USA}:
  {Association for Computing Machinery}, Jul. 2019, pp. 723--731.

\bibitem{yingHierarchicalGraphRepresentation2018}
Z.~Ying, J.~You, C.~Morris, X.~Ren, W.~Hamilton, and J.~Leskovec,
  ``Hierarchical {{Graph Representation Learning}} with {{Differentiable
  Pooling}},'' in \emph{Advances in {{Neural Information Processing Systems}}},
  vol.~31.\hskip 1em plus 0.5em minus 0.4em\relax {Curran Associates, Inc.},
  2018.

\bibitem{bianchiSpectralClusteringGraph2020}
F.~M. Bianchi, D.~Grattarola, and C.~Alippi, ``Spectral {{Clustering}} with
  {{Graph Neural Networks}} for {{Graph Pooling}},'' in \emph{Proceedings of
  the 37th {{International Conference}} on {{Machine Learning}}}.\hskip 1em
  plus 0.5em minus 0.4em\relax {PMLR}, Nov. 2020, pp. 874--883.

\bibitem{kipfSemiSupervisedClassificationGraph2017}
T.~N. Kipf and M.~Welling, ``Semi-{{Supervised Classification}} with {{Graph
  Convolutional Networks}},'' in \emph{{{ICLR}}}, Feb. 2017.

\bibitem{choiGraphNeuralControlled2022}
J.~Choi, H.~Choi, J.~Hwang, and N.~Park, ``Graph neural controlled differential
  equations for traffic forecasting,'' in \emph{Proceedings of the {{AAAI
  Conference}} on {{Artificial Intelligence}}}, vol.~36, Jun. 2022, pp.
  6367--6374.

\bibitem{yanLearningDynamicHierarchical2022}
H.~Yan, X.~Ma, and Z.~Pu, ``Learning {{Dynamic}} and {{Hierarchical Traffic
  Spatiotemporal Features With Transformer}},'' \emph{IEEE Transactions on
  Intelligent Transportation Systems}, vol.~23, no.~11, pp. 22\,386--22\,399,
  Nov. 2022.

\end{thebibliography}
\end{document}